\documentclass[sigconf]{acmart}
\usepackage{amsmath}
\usepackage{mathtools}
\usepackage{amsthm}
\usepackage{siunitx}
\usepackage{microtype}
\usepackage{graphicx}
\usepackage{subfigure}
\usepackage{booktabs} 
\usepackage{multirow}
\usepackage{multicol}
\usepackage{color}
\usepackage{colortbl}
\usepackage{bm}
\usepackage[super]{nth}
\usepackage[capitalize,noabbrev]{cleveref}
\definecolor{myMagenta}{HTML}{b800ae}
\definecolor{myBlue}{HTML}{3593f6}
\definecolor{lm_purple_low}{RGB}{240,240,248}
\definecolor{lm_purple}{RGB}{227,227,240}
\definecolor{lm_red}{RGB}{230,36,43}
\definecolor{cblue}{rgb}{0.21,0.49,0.74}
\AtBeginDocument{%
  }

\copyrightyear{2025}
\acmYear{2025}
\setcopyright{cc}
\setcctype{by}
\acmConference[KDD '25]{Proceedings of the 31st ACM SIGKDD Conference on
Knowledge Discovery and Data Mining V.2}{August 3--7, 2025}{Toronto, ON,
Canada}
\acmBooktitle{Proceedings of the 31st ACM SIGKDD Conference on Knowledge
Discovery and Data Mining V.2 (KDD '25), August 3--7, 2025, Toronto, ON,
Canada}
\acmDOI{10.1145/3711896.3736931}
\acmISBN{979-8-4007-1454-2/2025/08}

\newcommand{\ddg}{\Delta \Delta G}
\newcommand{\dde}{\Delta \Delta E}
\newcommand{\prot}[1]{\text{#1}}
\newcommand{\mut}{\text{mut}}
\newcommand{\wt}{\text{wt}}

\newcommand{\kbt}{k_{\text{B}}T}

\begin{document}

\title{Energy-Based Models for Predicting \\
Mutational Effects on Proteins}
\author{Patrick Soga}
\email{zqe3cg@virginia.edu}
\affiliation{%
  \institution{University of Virginia}
  \city{Charlottesville}
  \state{Virginia}
  \country{USA}
}

\author{Zhenyu Lei}
\email{vjd5zr@virginia.edu}
\affiliation{%
  \institution{University of Virginia}
  \city{Charlottesville}
  \state{Virginia}
  \country{USA}
}

\author{Yinhan He}
\email{nee7ne@virginia.edu}
\affiliation{%
  \institution{University of Virginia}
  \city{Charlottesville}
  \state{Virginia}
  \country{USA}
}

\author{Camille Bilodeau}
\email{cur5wz@virginia.edu}
\affiliation{%
  \institution{University of Virginia}
  \city{Charlottesville}
  \state{Virginia}
  \country{USA}
}

\author{Jundong Li}
\email{jundong@virginia.edu}
\affiliation{%
  \institution{University of Virginia}
  \city{Charlottesville}
  \state{Virginia}
  \country{USA}
}

\renewcommand{\shortauthors}{Patrick Soga, Zhenyu Lei, Yinhan He, Camille Bilodeau, and Jundong Li}

\begin{abstract}
    Predicting changes in binding free energy ($\Delta\Delta G$) is a vital task in protein engineering and protein-protein interaction (PPI) engineering for drug discovery. Previous works have observed a high correlation between $\Delta\Delta G$ and entropy, using probabilities of biologically important objects such as side chain angles and residue identities to estimate $\Delta\Delta G$. However, estimating the full conformational distribution of a protein complex is generally considered intractable.  In this work, we propose a new approach to $\Delta\Delta G$ prediction that avoids this issue by instead leveraging energy-based models for estimating the probability of a complex's conformation. Specifically, we novelly decompose $\Delta\Delta G$ into a sequence-based component estimated by an inverse folding model and a structure-based component estimated by an energy model. This decomposition is made tractable by assuming equilibrium between the bound and unbound states, allowing us to simplify the estimation of degeneracies associated with each state. Unlike previous deep learning-based methods, our method incorporates an energy-based physical inductive bias by connecting the often-used sequence log-odds ratio-based approach to $\Delta\Delta G$ prediction with a new $\Delta\Delta E$ term grounded in statistical mechanics. We demonstrate superiority over existing state-of-the-art structure and sequence-based deep learning methods in $\Delta\Delta G$ prediction and antibody optimization against SARS-CoV-2.\footnote{The authors wish to clarify that the performance differences between the final version of this work and the original submission arose due to implementation differences with~\citet{jiao2024boltzmann}, as the relevant code was unavailable at the time of submission.}
\end{abstract}



\maketitle

\section{Introduction}
Accurately modeling protein-protein interactions (PPIs) is a central task in therapeutic development and drug discovery. For example, antibody-antigen interactions govern crucial biological processes relating to combating infectious diseases~\citep{schoeder2021modeling,ruffolo2023fast}, and nanobodies targeting checkpoints such as PD-1 show promise in cancer immunotherapy~\citep{Darvin2018ImmuneCI}. Central to characterizing the behavior of a given protein is its binding free energy with its target, or $\Delta G$, which directly influences important properties such as stability~\citep{kastritis2012daring,du2016insights}. An important task in protein design involves finding favorable mutations to residues that decrease the $\Delta G$ of a PPI, leading researchers to study factors that influence changes in binding free energy ($\ddg$). This is difficult given the intractably large space of possible mutations, resulting in increased demand for computational methods to estimate mutational effects on proteins without the need to synthesize and validate mutations in a wet lab~\citep{marchand2022computational}.

Traditional methods for predicting $\ddg$ use techniques such as empirical energy functions \& force fields~\citep{alford2017rosetta,delgado2019foldx,park2016simultaneous} and BLOSUM matrices~\citep{Montanucci2019-uu,Henikoff1992-qu}. Recently, however, deep learning techniques have shown promise, outperforming classical methods~\citep{liu2023predicting,luo2023rotamer,shan2022ddgpred,jiao2024boltzmann}. While classical methods ground $\ddg$ in hand-crafted features and domain-expert-guided heuristics, deep learning approaches instead leverage the connection between the configurational entropy of a protein and its energy landscape, commonly using generative pre-training tasks as a means of approximating the true $\ddg$ of a given PPI. While previous methods~\citep{liu2023predicting,luo2023rotamer,meier2021language} have found success in using learned side chain and residue sequence probabilities as a proxy for binding free energy, they typically assume that the underlying backbone structure of the protein is not perturbed under mutation. As~\citet{mo2024multi} point out, models which do not take backbone modeling into account typically underperform on mutations between glycine and proline, residues with completely different levels of backbone flexibility. Similarly, tools such as FoldX~\citep{delgado2019foldx} that do not move backbones during inference have been shown to introduce bias in their $\ddg$ predictions and end up needing extensive corrective backbone relaxation~\citep{Usmanova2018-qn}. This suggests that $\ddg$ prediction crucially depends on allowing for flexibility in protein structure. Furthermore, while previous classically-based methods~\citep{Sheng2023-gd,Cui2020-ji} are often grounded in domain-expert guidance, deep learning approaches to $\ddg$ prediction typically lack a strong inductive bias that is consistent with physical theory. While~\citet{jiao2024boltzmann} and~\citet{dutton2024improving} are among the first to incorporate unbound state log-probabilities with inverse folding models inspired by the thermodynamic cycle, they use the same unrealistic fixed-backbone assumption as previous works, leaving a research gap for the development of deep learning-based models with strong physical inductive biases for $\ddg$ prediction.

\noindent \textbf{This Work.} In this work, we propose a new strategy for estimating $\ddg$ prediction rooted in statistical mechanics by leveraging the relationship between a protein's free energy and the probability of its conformation. We first rethink the thermodynamic definition of $\ddg$ and follow the derivation used by~\citet{jiao2024boltzmann} to motivate the use of inverse folding models typically seen in the existing $\ddg$ literature. We then relax the commonly held fixed-backbone assumption in previous deep learning-based methods and show that this results in the emergence of a $\dde$ term representing the change in energy under a set of mutations. Unique to our analysis is the consideration of the canonical ensemble of a given PPI, which we show allows us to sample only a single mutant structure, thereby making the estimation of the $\dde$ term tractable. We then estimate this $\dde$ term using a diffusion-based score-matching model following~\citet{jin2023dsmbind} to both estimate mutated residue coordinates and predict the change in binding free energy given these mutations. We show that our approach, dubbed \textbf{EBM-DDG} (\underline{E}nergy-\underline{B}ased \underline{M}odel-DDG), outperforms current state-of-the-art models on the SKEMPI v2.0 dataset~\citep{Jankauskaite2019-vi} in $\ddg$ prediction and ranking CDR mutations for neutralization against SARS-CoV-2~\citet{shan2022ddgpred}. Specifically, our contributions are as follows:
\begin{itemize}
    \item We propose a new analysis of the thermodynamic definition of $\ddg$ in the preceding literature based on statistical mechanics that removes the fixed backbone assumption and makes energy computation tractable. To the best of our knowledge, our method is the first in the deep learning-based literature to remove this assumption by means of energy-based models.
    \item We incorporate a corrective energy term that can account for changes in the backbone caused by mutations, providing a stronger inductive bias for $\ddg$ prediction.
    \item We demonstrate that our proposed approach EBM-DDG outperforms state-of-the-art methods in $\ddg$ prediction and CDR mutation ranking for SARS-CoV-2.
\end{itemize}

\section{Related Work}
Traditional $\ddg$ prediction methods can be briefly categorized as biophysics-based or statistical-based. Biophysics-based methods involve empirical energy functions derived from phenomena such as van der Waals forces and electrostatic potentials~\citep{alford2017rosetta,Barlow2018-qx,delgado2019foldx} while statistical-based methods focus on engineering features such as BLOSUM matrices~\citep{Montanucci2019-uu,Henikoff1992-qu} and evolutionary conservation scores~\citep{li2016mutabind}. These traditional methods are limited by their limited features, highlighting the need for more powerful approaches to $\ddg$ prediction. Deep learning-based approaches combat this shortcoming, categorizable as sequence-based or structure-based. Sequence-based methods such as ESM-1v~\citep{meier2021language}, MIF~\citep{Yang2023-ds}, and Boltzmann Alignment~\citep{jiao2024boltzmann} use pre-trained protein language models or inverse folding models to learn residue probabilities and use them to predict $\ddg$, the idea being that the predicted $\ddg$ should correspond to the difference between the predicted log-likelihood of the mutant sequence and that of the wild-type sequence. Structure-based methods rely on a similar insight except that they use the predicted likelihood of the mutant/wild-type structures of the PPI rather than its sequence. RDE-Network~\citep{luo2023rotamer} learns a normalizing flow to estimate side chain probability distributions as a pre-training task and then finetunes the resulting embeddings for $\ddg$ prediction. DiffAffinity~\citep{liu2023predicting} uses the same finetuning approach but opts to pre-train a denoising diffusion model rather than a normalizing flow to learn side chain angles. ProMIM~\citep{mo2024multi} leverages multiscale modeling, pre-training a 3D encoder to predict positive pairs of binding proteins, contact maps between chains, and side-chain angles. Prompt-DDG~\citep{wu2024prompt} learns prompt features by autoencoding each residue's microenvironment using a VQ-VAE~\citep{oord2017neural} and uses the resulting codebook embeddings for downstream $\ddg$ prediction. Finally, Refine-PPI~\citep{wu2025dynamicsinspired} models interface residues with 3D Gaussians for estimating complex probabilities and sampling their coordinates to compute $\ddg$.  Our method falls at the intersection of these sequence- and structure-based methods, building on the sequence-based advancements of~\citet{jiao2024boltzmann} and incorporating an energy term that accounts for the probability of the structure of the whole complex, thereby providing a more comprehensive inductive bias.

\section{Preliminaries}
\subsection{Notation}
A \textit{protein chain} A is a sequence of $n$ amino acids (also known as \textit{residues}) $S_\prot{A} \in \{1, \dots, 20\}^n$ with 3D Cartesian coordinates $X_{\prot A} \in \mathbb R^{n \times 3}$. A \textit{protein complex} AB is simply the union of two individual protein chains A and B that can be considered to be in a ``\textit{bound}'' or ``\textit{unbound}'' state $\prot{AB}_b$ or $\prot{AB}_u$. It is straightforward to generalize this notation to complexes with more than two chains. \textit{Residue mutations}, or simply mutations, in a chain A are tuples $(r, c, i, s)$ that stand for the replacement of residue $r$ in chain $c$ at index $i$ with residue $s$ in A. For example, $(3, 2, 442, 18)$ refers to mutating the \nth{442} residue of the \nth{2} chain in A from 3 to 18. A protein chain without any mutations is known as a \textit{wild-type} chain~\citep{Ruthner_Batista2005-nw}.

\subsection{Problem Definition -- $\ddg$}
Before defining $\ddg$, we first define $\Delta G$, or the \textit{binding free energy}, of a protein complex AB as 
\begin{equation}
    \label{eq:dg}
    \Delta G = G_b - G_u= -\kbt\left(\ln p(\text{AB}_b) - \ln p(\text{AB}_u)\right)
\end{equation}
where $k_{\text{B}}$ is the Boltzmann constant, $T$ is the thermodynamic temperature, and $p(\text{AB}_b)$ and $p(\text{AB}_u)$ stand for the probabilities of AB being in a bound or unbound state. $\ddg$, the change in binding free energy, is then defined as
\begin{equation}
    \label{eq:ddg}
    \Delta \Delta G = \Delta G^{\text{mut}} - \Delta G^{\text{wt}}
\end{equation}
where $\Delta G^{\text{mut}}$ and $\Delta G^{\text{wt}}$ stand for the binding free energies of the mutant and wild-type versions of AB. Our main task is to predict the $\ddg$ of a complex given a set of mutations. We are now ready to introduce prior work on the connection between $\ddg$ prediction and inverse folding that will serve as the foundation of our method.

\subsection{$\ddg$ Prediction Using Inverse Folding}
Inverse folding models are models that are trained to predict a protein's amino acid sequence given its 3D structure, estimating $p(\text{AB}) \approx p\left(S_\prot{AB} \mid X_\prot{AB}\right)$. Empirically, it is well known that there is a high correlation between the log-likelihood of sequences from these inverse folding models and binding energy~\citep{meier2021language,widatalla2024aligning,bennett2023improving}. Accordingly, many previous methods~\citep{meier2021language,bushuiev2024learning,Cagiada2024pred,widatalla2024aligning} use the log-odds ratio
$\ln p(S_{\text{AB}}^\text{mut} | X_{\text{AB}}) - \ln p(S_{\text{AB}}^\text{wt} | X_\text{AB}))$ to estimate $\ddg$.~\citet{jiao2024boltzmann} elucidate the connection between this ratio and $\ddg$. They first express $p(\prot{AB}) = p(X_\prot{AB} | S_\prot{AB})$ since we are assured at least the sequence of a given protein complex. Then, applying Baye's theorem, they rewrite \cref{eq:dg} to obtain
\begin{align}
        \Delta G &= -\kbt \left( \ln\frac{p\left(S_\prot{AB} | X_{b}\right) p\left(X_{b}\right)}{p\left(S_\prot{AB}\right)} - \ln \frac{p\left(S_\prot{AB}| X_u\right) p(X_{u})}{p\left(S_\prot{AB}\right)} \right),
\end{align}
resulting in the following formulation of $\ddg$:
\begin{align}
    &\ddg = \Delta G^\mut - \Delta G^\wt \nonumber \\ 
    &= -\kbt\left(\ln \frac{p\left(S_\prot{AB}^\mut | X_{b}^\mut\right) p\left(X_{b}^\mut\right)}{p\left(S_\prot{AB}^\mut| X_{u}^\mut\right) p\left(X_{u}^\mut\right)}
     - \ln \frac{p\left(S_\prot{AB}^\wt | X_{b}^\wt\right) p\left(X_{b}^\wt\right)}{p\left(S_\prot{AB}^\wt| X_{u}^\wt\right) p\left(X_{u}^\wt\right)}\right). 
    \label{eq:full-ddg}
\end{align}
Crucially,~\citet{jiao2024boltzmann} then assume that the 3D structure of the protein complex is \textbf{unchanged} after mutation, namely $X^\mut = X^\wt$. This allows the unconditioned $p\left(X\right)$ terms to cancel, resulting in the core formulation for the authors' method dubbed BA-DDG (\underline{B}oltzmann \underline{A}lignment-DDG):
\begin{equation}
    \label{eq:boltzmann}
    \ddg_\text{BA} = -\kbt \left(\ln \frac{p\left(S_\prot{AB}^\mut | X_b^\mut\right)}{p\left(S_\prot{AB}^\mut | X_u^\mut\right)} - \ln \frac{p\left(S_\prot{AB}^\wt | X_b^\wt\right)}{p\left(S_\prot{AB}^\wt | X_u^\wt\right)} \right)
\end{equation}
where $p(S_\prot{AB}^\mut | X_u^\mut)$ is approximated with the product
$p(S_\prot{A}^\mut |  X_\prot{A}^\mut) \cdot p(S_\prot{B}^\mut | X_\prot{B}^\mut)$ by treating unbound protein chains as being relatively far apart and therefore interacting minimally and independently. While $X_b^\mut = X_b^\wt$ under this assumption, we keep their notation distinct for consistency with the previous work. In summary, $\ddg_{\text{BA}}$ is essentially a corrected version of the log-odds ratio used in previous works that either ignored the unbound states of each protein chain~\citep{meier2021language,hsu2022esm1f} or treated the unbound states as single residues~\citep{dutton2024improving}. BA-DDG is a strong performer, outperforming the state-of-the-art at the time of release by up to 16\% in per-complex Pearson correlation coefficient. However, although assuming $X^\mut = X^\wt$ may be reasonable for mutations that do not have a significant effect on the structure of a complex, the effects of mutations, such as mutating proline to glycine and vice versa~\citep{zhou2024ddmut}, that may have a substantial effect on structure may be inaccurately predicted. Such an assumption is also a violation of physical principles seeing as mutations often make subtle structural modifications which may affect $\ddg$~\citep{Naganathan2010-kq}. Such a simplification may seem inevitable since the space of possible conformations for proteins is so vast. Our method aims to overcome this challenge by leveraging energy models to tractably estimate conformation probabilities, forming a stronger inductive bias for $\ddg$ prediction.

\section{Our Method}
We now begin discussion of our core method. We first introduce an analysis of~\cref{eq:full-ddg} from the perspective of energy-based models followed by our key insight involving the equilibrium of the bound and unbound structures that makes our method tractable. We then elaborate on the full details of our model and our final loss function before proceeding to our experiments.

\subsection{Relaxing the Fixed Structure Assumption}
\begin{figure}[!htbp]
    \centering
    \includegraphics[width=\linewidth]{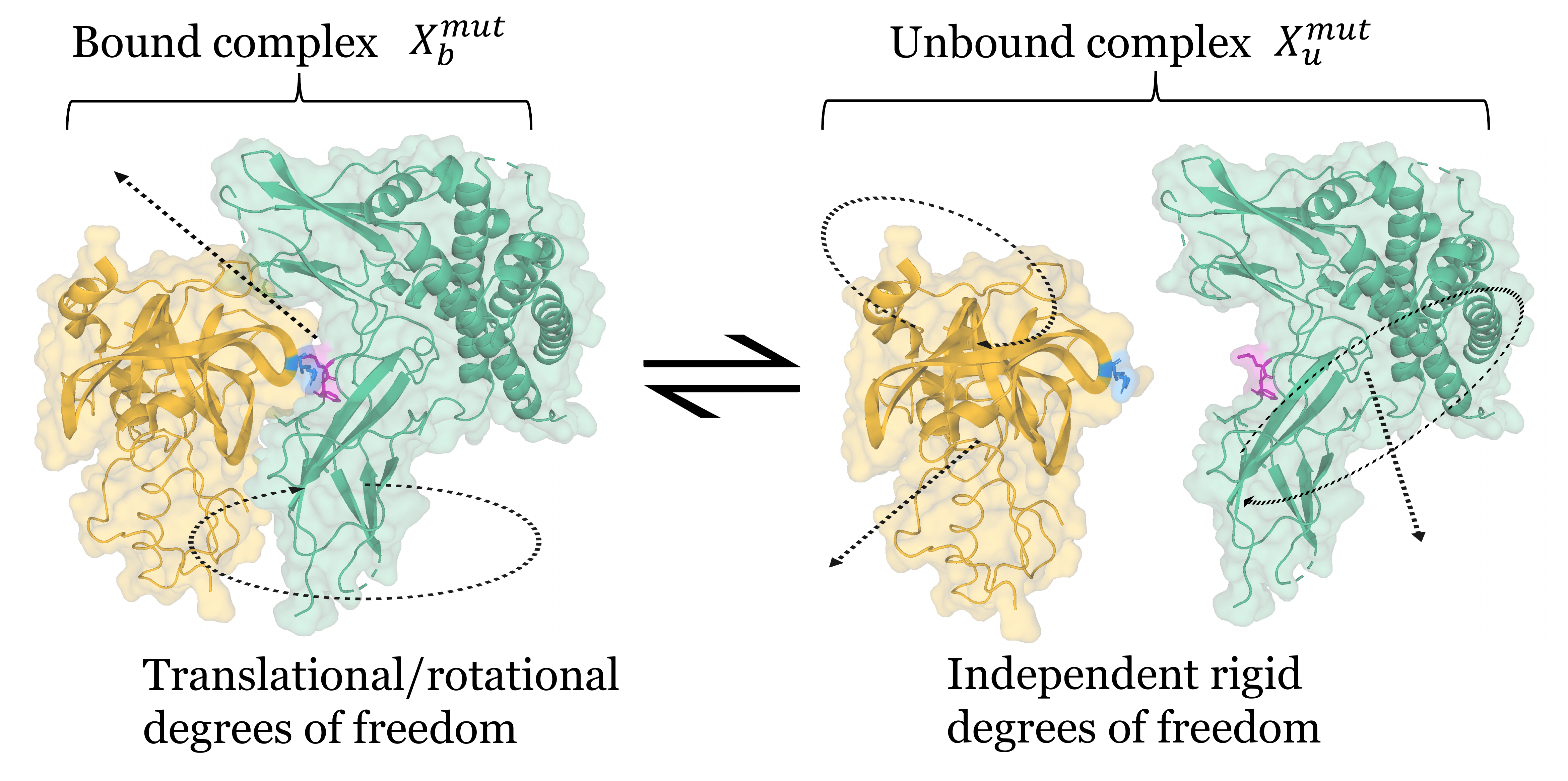}
    \caption{Illustration of the canonical ensemble for the mutant structure $X^\mut$. An identical comparison can be made for the wild-type structure. The $\Omega$ degeneracy terms for the bound/unbound states are associated with the rotational and translational degrees of freedom of the complex/chains, respectively. The ratio of microstates is driven by geometry rather than sequence, allowing us to simplify~\cref{eq:ddg-ours-full}.}
    \Description[Canonical ensemble for a PPI considers the PPI exchanging between bound and unbound states]{We consider the bound and unbound states of a PPI and their degeneracy terms.}
    \label{fig:ensemble}
\end{figure}
Our key insight is to recognize that we can relax the fixed structure assumption by applying statistics arising from the canonical ensemble in statistical mechanics to incorporate the $p\left(X\right)$ terms. Specifically, we consider the 3D structure of $\prot{AB}$ in the canonical ensemble, or the space of possible configurations of the structure between its bound and unbound states, as visualized in~\cref{fig:ensemble}. The probability of the structure $X$ in the canonical ensemble can be expressed as:
\begin{equation*}
    p(X) = \frac{\Omega_\text{AB}\exp\left(-\beta E\left(X\right)\right)}{Z_{\prot{AB}}};\quad Z_{\prot{AB}} = \sum\limits_{X_i} \exp\left(-\beta {E}\left(X_i\right)\right)
\end{equation*}
where $\Omega_\prot{AB}$ refers to the the number of configurations (microstates) corresponding to the structure $X$, $\beta = 1/k_\text{B} T$, and $E(X)$ refers to the total energy of the system of $X$~\citep{mcquarrie2000statistical}.  The denominator, $Z_\prot{AB}$, is known as the \textit{partition function}, which serves as the normalizing constant that enforces the probability distribution to sum to unity. In physical terms, $Z_\prot{AB}$ is the sum over all Boltzmann factors $\exp(-\beta E\left(X_i\right))$ for all possible configurations of $\prot{AB}$. Importantly, because of this definition, $Z_\prot{AB}$ does not depend on whether the structure $X$ is bound or unbound, but rather encompasses all possible states corresponding to the protein complex $\prot{AB}$. In contrast, $\Omega_\prot{AB}$ does depend on whether $X_\prot{AB}$ is in the bound or unbound state because the number of microstates corresponding to these two states will differ. To this end, we will use $\Omega^{b}_\prot{AB}$ and $\Omega^{u}_{\prot{AB}}$ to denote the degeneracies of the bound and unbound states, respectively. Introducing these definitions into Equation (4) yields the following $\Delta G$ formulation:
\begin{equation}
    \label{eq:dg-energy}
    \Delta G = -\kbt\left(\ln \frac{p(S_\prot{AB} | X_{b}) \cdot \Omega^b_{\prot{AB}}\exp\left(-\beta{E(X_b)}\right) } {{p(S_\prot{AB}| X_{u})  \cdot \Omega^u_{\prot{AB}}\exp\left(-\beta E(X_u)\right)} }\right)
\end{equation}
where $E(X_b)$ and $E(X_u)$ represent the total energies associated with $\prot{AB}$ in the bound and unbound states, respectively, and can therefore be straightforwardly calculated using a neural network potential/energy function~\citep{Batatia2022mace,Musaelian2022-sf,jin2023dsmbind} or any classical protein force field. $\Omega_{b}$ and $\Omega_{u}$ are more challenging to calculate as they represent the number of microstates associated with the bound and unbound structures. Importantly, we note that while $Z_\prot{AB}$ is intractable to compute, it cancels when calculating changes in free energy, meaning we only need to sample a single representative mutant structure.

Using Equation (6), we can reformulate $\Delta\Delta G$ for a mutation from $\prot{AB}$ to $\prot{AB}_\mut$ as
\begin{align}
    \label{eq:ddg-ours-full}
    \ddg = &-\kbt \left(\ln \frac{p\left(S_\prot{AB}^\mut | X_b^\mut\right)}{p\left(S_\prot{AB}^\mut | X_u^\mut\right)} - \ln \frac{p\left(S_\prot{AB}^\wt | X_b^\wt\right)}{p\left(S_\prot{AB}^\wt | X_u^\wt\right)} \right) \nonumber \\
    &-\kbt \ln\left(\frac{\exp\left(-\beta E\left(X_b^\mut\right)\right)\exp\left(-\beta E\left(X_u^\wt\right)\right)}{\exp\left(-\beta E\left(X_u^\mut\right)\right)\exp\left(-\beta E\left(X_b^\wt\right)\right)}\right) \nonumber \\
    & -\kbt \ln\left(\frac{\Omega_b^{{\mut}}\Omega_u^\wt}{\Omega_u^{{\mut}}\Omega_b^\wt}\right).
 \end{align}
We now discuss our key insight allowing us to cancel the final degeneracy term in~\cref{eq:ddg-ours-full}. Note that $\Omega_b$ and $\Omega_u$ represent the number of microstates associated with the structures $X_b$ and $X_u$. Since $X_b$ is a static configuration, the only degrees of freedom that allow for multiple microstates come from rotating $X_b$ about its principal components and allowing it translational degrees of freedom. The primary differences between the degrees of freedom of the bound state $X_b$ and those of the unbound state $X_u$ come from the fact that, in the unbound state, each individual protein can rotate and translate independently. Hence, as long as we consider $\prot{AB}$ to be well-represented by a static structure, the ratio of microstates ${\Omega_b}/{\Omega_u}$ is \textit{driven by geometry} rather than protein sequence. Hence, residue mutations from $\prot {AB}$ to $\prot{AB}_\mut$ will not significantly affect the difference in the ratios $\Omega_b^{\mut}/\Omega_u^{\mut}$ and ${\Omega_b^\wt}/{\Omega_u^\wt}$, leading us to conclude that
\begin{equation}
    \label{eq:omegas}
    \frac{\Omega_b^{{\mut}}}{\Omega_u^{{\mut}}} = \frac{\Omega_b^\wt}{\Omega_u^\wt}.
\end{equation}
We expect this assumption to hold for most PPIs, such as those present in SKEMPI~\citep{Jankauskaite2019-vi}, since they can be crystallized and experimentally observed in stable, static 3D structures. In contrast, proteins such as intrinsically disordered proteins~\citep{Tompa2002-hm,Wright2015-ja} or membrane proteins~\citep{Carpenter2008-kk} do not adopt a single well-defined structure, meaning changes in residues may introduce significant new degrees of freedom, resulting in extensive structural ensembles that break our assumption of a static $X$ and produce disparate microstate ratios.

We reiterate that, in contrast with prior work by~\citep{jiao2024boltzmann}, our approach does not require that $X^\wt = X^\mut$, but only that we can identify a valid representative structure $X$ for a sequence $S_\prot{AB}$, which is a much more realistic assumption when working with PPIs that are stably representable with a static structure. Leveraging this assumption, we can use~\cref{eq:omegas} to simplify \cref{eq:ddg-ours-full} and arrive at our core $\ddg$ formulation:
\begin{align}
    \label{eq:ddg-ours}
    \ddg &= -\kbt \ln \left(\frac{p\left(S_{\prot{AB}}^{\mut} | X_b^\mut\right) p\left(S_\prot{AB}^\wt | X_u^\wt\right)}{p\left(S_{\prot{AB}}^{\mut} | X^\mut_u\right) p\left(S_\prot{AB}^\wt | X^\wt_b\right)}\right) - \Delta \Delta E \nonumber \\
    &= \ddg_{\text{BA}} - \dde
\end{align}
where 
\begin{align*}
    \Delta \Delta E &= \Delta E_{\prot{AB}_{\mut}}-\Delta E_\prot{AB} \\
    &= \left(E\left(X_b^\mut\right)-E\left(X_u^\mut\right)\right)-\left(\left(E(X_b^\wt\right)-E\left(X_u^\wt\right)\right).
\end{align*}

In summary, we have relaxed the previous fixed structure assumption by introducing a simple term,  $\Delta \Delta E$, which can be easily calculated from an energy model or a classical protein force field. Assuming the wild-type complex structure is available, we only need to sample a single representative structure for $X^\mut$ and apply our energy model to compute $\dde$, circumventing the need to enumerate all Boltzmann factors or compute any degeneracy terms. To summarize, our analysis decomposes $\ddg$ into (1) an inverse folding component that leverages a sequence-based log-odds ratio correlated with the difference in binding energy and (2) a corrective energy term based on a sampled structure in a local minimum of the PPI's energy surface. Intuitively, (1) acts as a coarse approximation of $\ddg$ by capturing how mutations shift residue preferences, while (2) refines this by taking into account conformational differences between the mutant and wild-type complexes. Together, these terms form the basis of our method for $\ddg$ prediction with an inductive bias grounded in statistical mechanics.
\subsection{Incorporating Energy for $\ddg$}
\begin{figure*}[!htbp]
    \centering
    \includegraphics[scale=0.10]{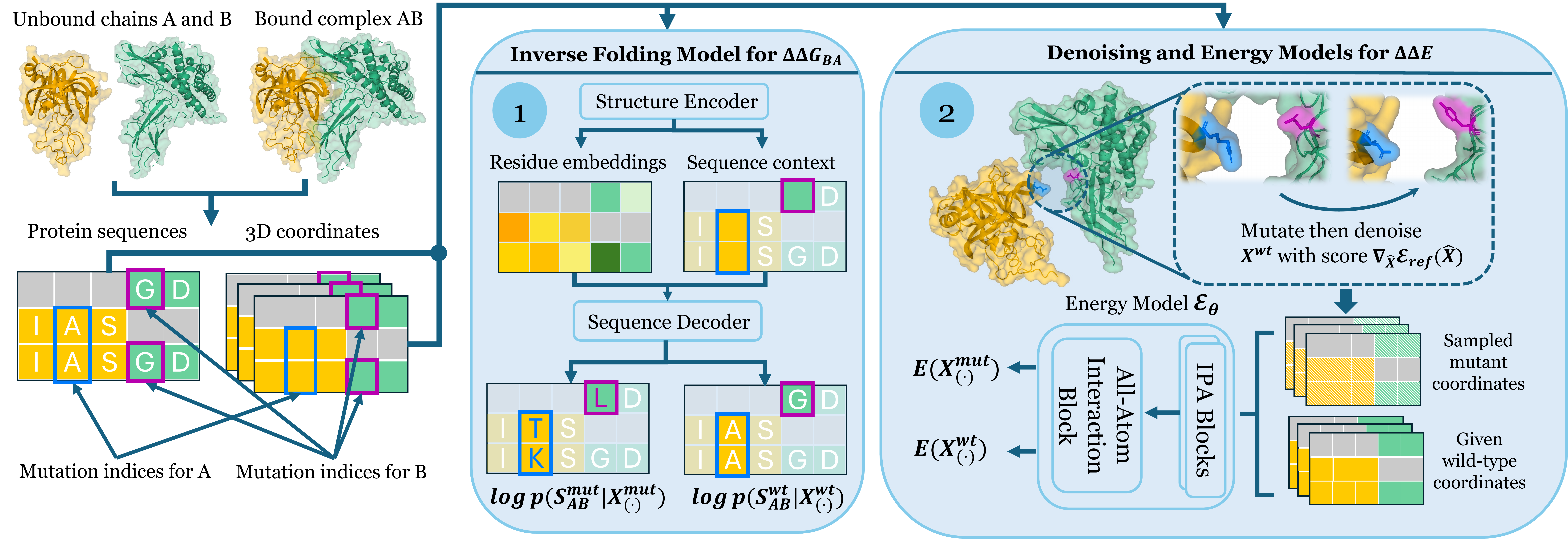}
    \caption{Visualization of our pipeline for EBM-DDG. From left-to-right, we bundle the chains A and B as well as the complex AB in a single batch where gray boxes stand for masked positions. The \textcolor{myBlue}{blue} and \textcolor{myMagenta}{purple} highlighted residues and their respective coordinates represent indices of a given mutation for chains A and B. First, wild-type and mutant sequences $S^\wt_\prot{AB}$ and $S^\mut_{\prot{AB}}$ of the complex and its chains as well as its wild-type crystal structure $X^\wt_\prot{AB}$ are passed into an inverse folding model where each mutant residue's log probabilities are used to compute the $\ddg_\text{BA}$ term. We then pass the wild-type sequence and structure into a reference EBM $\mathcal{E}_\text{ref}$ whose gradient we use to estimate the geometry of the mutant residues $\hat{X}_\prot{AB}^\mut$ and use a trainable EBM $\mathcal E_\theta$ composed of a shallow Invariant Point Attention network followed by an all-atom interaction block to predict the wild-type and mutant $\dde$ terms in \cref{eq:ddg-ours}. Subscripts $(\cdot)$ are placeholders for both bound and unbound states of the PPI.}
    \Description[Overall pipeline]{Every protein complex and its constituent chains are fed into a trainable inverse folding model for sequence log-probability calculation. Their structures are fed into a reference energy function for sampling of mutated coordinates. These coordinates are then fed to a trainable energy function to evaluate the difference in energy. Both sequence and structure terms are combined for final DDG prediction.}
    \label{fig:pipeline}
\end{figure*}
So far, we have shown how to eliminate the need to approximate partition functions and degeneracy terms and therefore the need to sample a variety of structures by considering the canonical ensemble of $X$. We now discuss how to leverage \cref{eq:ddg-ours} in detail. 
\subsubsection{Choice of Energy Model.}
$\ddg_{\text{BA}}$ can be easily computed using inverse folding models. We use ProteinMPNN~\citep{dauparas2022robust} as our backbone inverse folding model and retain the approximation of unbound chains $p(S_\prot{AB}^\mut \mid X_u^\mut) \approx p(S_\prot{A}^\mut \mid X_\prot{A}^\mut) \cdot p(S_\prot{B}^\mut \mid X_\prot{B}^\mut)$ following~\citet{jiao2024boltzmann}. We leave a comprehensive treatment of modeling unbound chains to future work. On the other hand, $\dde$ requires two components: a generative model $f$ from which we will sample a low-energy structure $\hat X \sim f$, and an energy function $\mathcal E$ that will evaluate the energy of $\hat X$. Preferably, we want to sample from a distribution defined by our generative model $p_f(X)$ that is well-aligned with $E$; i.e., we aim for $p_f(\hat X) \propto \exp(-\mathcal E(\hat X))$. To satisfy this criterion, we opt to follow DSMBind~\citep{jin2023dsmbind} and use a learned energy function $E$ whose gradient is used to learn the score $\nabla_{\hat X} \log p(\hat X)$, which can be interpreted as a conservative force field~\citep{arts2023two}. Specifically, DSMBind learns to match its score $\partial \mathcal E/\partial \hat{X}$ for a randomly rotated and translated $\hat X$ with the score of the corresponding translational noise $\nabla_t \log p(t)$ and rotational noise $\nabla_\omega \log p(\omega)$. Minimizing this objective is equivalent to learning a crystal structure at a local minimum of an EBM~\citep{zaidi2023pretraining}, which implies that samples from DSMBind will correspond to low energy (high likelihood) structures. With this, we can generate mutant structures of high likelihood $\hat X^\mut$, and then use the learned energy function to compute the terms involving $E(X^\mut)$ which we plug into \cref{eq:ddg-ours}. To this end, we sample structures according to Langevin dynamics using DSMBind's learned score function, which we detail in~\cref{sec:sampling}. We note that, for the wild-type terms, we do not need to sample anything since we are given the wild-type crystal structures in our datasets. While there are other strong denoising-based generative models available such as AlphaFold 3~\citep{Abramson2024-nv} and RFDiffusion~\citep{Watson2023-ff}, we note that such models are trained without an explicit energy function. Recovering $\mathcal E(x)$ from a learned score function $s_\theta \approx -\nabla_x \mathcal E(x)$ is challenging and remains out of scope of this work.

\subsubsection{Supervision.}
In our experiments, we let $\kbt$ be a learned constant. Our objective function is the simple MSE loss
\begin{equation}
    \mathcal L = \text{MSE}\left(\ddg, \widehat{\ddg}\right);
    \quad \widehat{\ddg} = \ddg_{\text{BA}} - \dde,
\end{equation}
where $\dde$ is computed using a pre-trained energy model $\mathcal E_\theta$ that utilizes the the candidate mutant structure generated by a frozen reference energy model $\mathcal E_{\text{ref}}$. This ensures that the sampled structures are not adversely affected by the process of learning to predict $\ddg$. We also note that the minima of the denoising score-matching objective are equal up to an affine transformation of $\mathcal E$, meaning that DSMBind alone cannot predict absolute $\ddg$ values. To ease optimization, we therefore incorporate learnable scaling and bias correction terms $s, b \in \mathbb R$ to the $\dde$ term. Then, we use a pre-trained ProteinMPNN as our inverse folding model $p_\theta$ to calculate $\ddg_\text{BA}$. We follow~\citet{jiao2024boltzmann} and add their auxiliary KL-divergence loss $\beta D_{\text{KL}}(p_\theta(S|X) || p_\text{ref}(S | X))$ to avoid catastrophic forgetting by maintaining the distribution of the original pre-trained model $p_\text{ref}$ with weight $\beta = 0.001$.
\section{Experiments}
We evaluate EBM-DDG on two tasks: $\ddg$ prediction on SKEMPI v2.0~\citep{Jankauskaite2019-vi} and antibody optimization against SARS-CoV-2~\citep{shan2022ddgpred}. We then conduct a parameter sensitivity study with respect to the number of denoising steps in~\cref{sec:antibody}. We conclude with a case study on mutations that affect backbone coordinates in~\cref{sec:flex}.
\subsection{Experimental Settings}
For inverse folding, we use a version of ProteinMPNN from~\citep{dauparas2022robust} pre-trained on soluble proteins with an embedding dimension of $256$. See~\cref{sec:proteinmpnn} for more implementation details for ProteinMPNN. For the energy model, due to CUDA implementation issues, we replaced the SRU++-based frame averaging encoder~\citep{lei2021srupp,puny2022frame} with a shallow all-atom Invariant Point Attention network~\citep{Jumper2021-hp} and pre-train the model from scratch on crystal structures from SKEMPI v2.0~\citep{Jankauskaite2019-vi} under the Gaussian denoising score-matching loss. Given that structural changes are often relatively local upon mutation, we focus on denoising the coordinates of only the mutated residues and the immediate residues neighboring them. On SKEMPI v2.0, we train our pipeline for 30K steps using the Adam optimizer with a learning rate of 1e-4 and a batch size of 2. We use such a small batch size since we require the sequences and structures of both complexes and their chains in a single batch of mutations. During inference, we denoise coordinates for 10 steps, and we perform a sensitivity analysis to the number of steps in~\cref{sec:T}. All experiments were conducted using one NVIDIA A100 80GB GPU on a server with a 24-core AMD EPYC 7473X CPU.
\subsection{$\ddg$ Prediction -- SKEMPI v2.0}
We first evaluate EBM-DDG on SKEMPI v2.0~\citep{Jankauskaite2019-vi}, the largest and most popular benchmark for $\ddg$ prediction in the deep learning literature. SKEMPI v2.0 contains 7,085 mutations across 348 protein complexes with their wild-type crystal structures and $\ddg$ labels.

\noindent \textbf{Baselines.}
We compare our method with a variety of unsupervised and supervised baselines using reported numbers whenever possible. Unsupervised methods encompass traditional techniques including the {Rosetta} energy function~\citep{alford2017rosetta} and {FoldX}~\citep{delgado2019foldx}. Next, we compare with sequence- and evolutionary-based methods including {ESM-1v}~\citep{meier2021language}, Position-Specific Scoring Matrix ({PSSM}), the {MSA Transformer}~\citep{rao2021msa}, and {Tranception}~\citep{notin2022tranception}. Next, we have unsupervised pre-trained methods including a model pre-trained to predict residue {B-factor}s~\citep{luo2023rotamer}, {ESM-IF}~\citep{hsu2022esm1f}, {MIF-$\Delta$logit}~\citep{Yang2020-ur}, and {RDE-Linear}~\citep{luo2023rotamer}, and {BA-Cycle}~\citep{jiao2024boltzmann}. Finally, the strongest baselines are supervised models including {DDGPred}~\citep{shan2022ddgpred}, an end-to-end RDE-MLP hybrid model by~\citep{luo2023rotamer} (denoted by {End-To-End}), {MIF-Network}~\citep{Yang2020-ur}, {RDE-Network}~\citep{luo2023rotamer}, {DiffAffinity}~\citep{liu2023predicting}, {Prompt-DDG}~\citep{wu2024prompt}, {ProMIM}~\citep{mo2024multi}, and {BA-DDG}~\citep{jiao2024boltzmann}.
\begin{figure*}[!htbp]
    \centering
    \includegraphics[width=0.8\linewidth]{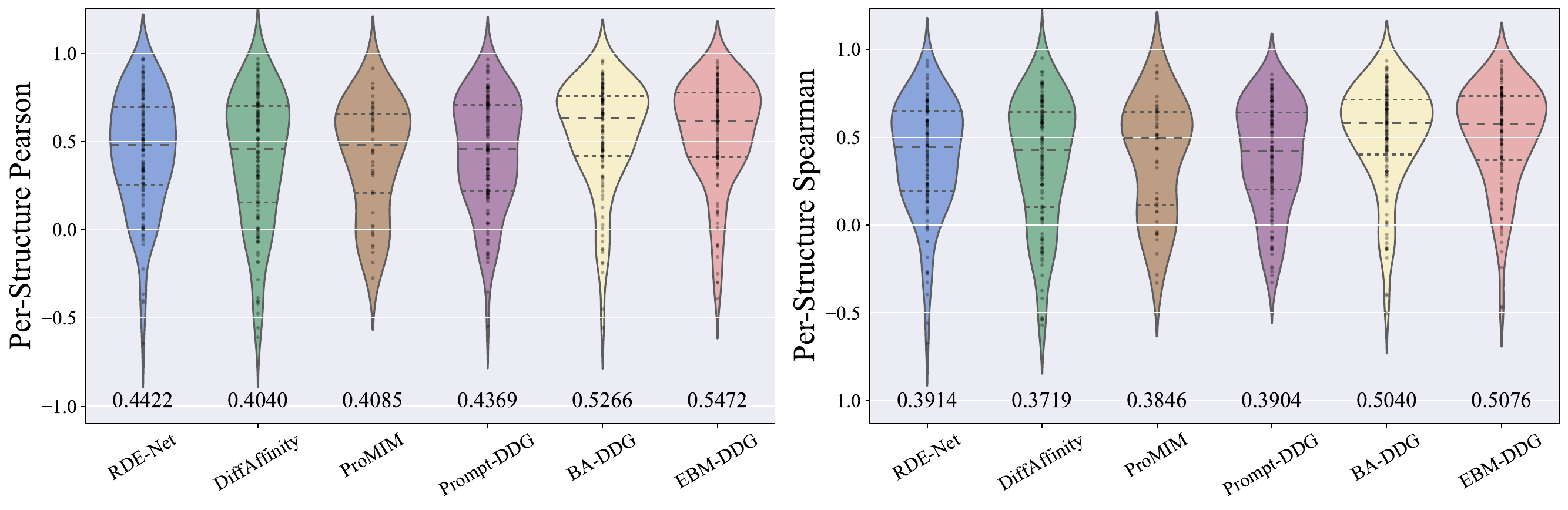}
    \caption{Test per-structure Pearson correlation and Spearman correlation score among the strongest baselines and EBM-DDG.}
    \Description[Violin plots for per-structure correlation performance]{EBM-DDG provides stronger mean per-structure performance as well as lower variance.}
    \label{fig:corr}
\end{figure*}

\noindent \textbf{Evaluation Metrics.}
We evaluate $\ddg$ prediction performance using 5 metrics: Pearson correlation coefficient, Spearman correlation coefficient, root mean squared error (RMSE), mean absolute error (MAE), and AUROC. AUROC for $\ddg$ prediction measures classification performance where labels correspond to the sign of the ground-truth $\ddg$ values. Practically, correlations of mutations within specific complexes is often more important practically when a specific antibody or protein is being studied~\citep{wu2024prompt,jiao2024boltzmann}. Therefore, we also report the mean Pearson/Spearman correlation coefficients for individual structures in addition to the mean correlation coefficients for all structures in aggregate. Following~\citep{wu2024prompt,luo2023rotamer}, we perform 3-fold cross-validation where each fold contains unique protein complexes so that each mutation is tested at least once.

\noindent\textbf{Setting with Dedicated Test Set.} During implementation, we noticed that the cross-validation code used by many baselines only used training/validation splits. From this, it was unclear what data test metrics were reported and on what data each model was tuned on. For example, the authors of Prompt-DDG clarified that they followed previous works and reported mean results across validation folds only without a test set. Therefore, for added rigor and to test for robustness, we set aside 10\% of each training fold to use as a validation set and report the mean metrics over the test folds rather than the validation folds. We do the same for RDE-Network, DiffAffinity, Prompt-DDG, and ProMIM since they all share the same dataset splitting code. For these models and our model, the first row in~\cref{tab:1} corresponds to metrics under the original setting, and the second corresponds to metrics under our new setting to more robustly compare models.
\begin{table*}[!htbp]
\begin{center}
\vspace{-1em}
\caption{3-fold cross-validation results on the SKEMPI v2.0. dataset where \textbf{bold} and \underline{underline} denote the best and second metrics. For models which we benchmark under our train/val/split setting, there may be two bold and underlined metrics.}
\label{tab:1}
\resizebox{0.85\textwidth}{!}{
\begin{tabular}{llccccccc}
\toprule
\multirow{2}{*}{\textbf{Category}} & \multirow{2}{*}{\textbf{Method}} & \multicolumn{2}{c}{\textbf{Per-Structure}} & \multicolumn{5}{c}{\textbf{Overall}} \\ \cmidrule(r){3-4} \cmidrule(r){5-9}
 &  & \textbf{Pearson} $\uparrow$ & \textbf{Spear.} $\uparrow$ & \textbf{Pearson} $\uparrow$ & \textbf{Spear.} $\uparrow$ & \textbf{RMSE} $\downarrow$ & \textbf{MAE} $\downarrow$ & \textbf{AUROC} $\uparrow$ \\ \midrule
\multirow{2}{*}{Energy Function} & Rosetta & 0.3284 & 0.2988 & 0.3113 & 0.3468 & 1.6173 & 1.1311 & 0.6562 \\
 & FoldX & 0.3789 & 0.3693 & 0.3120 & 0.4071 & 1.9080 & 1.3089 & 0.6582 \\ \midrule
\multirow{4}{*}{Sequence-based} & ESM-1v & 0.0073 & -0.0118 & 0.1921 & 0.1572 & 1.9609 & 1.3683 & 0.5414 \\
 & PSSM & 0.0826 & 0.0822 & 0.0159 & 0.0666 & 1.9978 & 1.3895 & 0.5260 \\
 & MSA Transformer & 0.1031 & 0.0868 & 0.1173 & 0.1313 & 1.9835 & 1.3816 & 0.5768 \\
 & Tranception & 0.1348 & 0.1236 & 0.1141 & 0.1402 & 2.0382 & 1.3883 & 0.5885 \\ \midrule
\multirow{4}{*}{Unsupervised} & B-factor & 0.2042 & 0.1686 & 0.2390 & 0.2625 & 2.0411 & 1.4402 & 0.6044 \\
 & ESM-1F & 0.2241 & 0.2019 & 0.3194 & 0.2806 & 1.8860 & 1.2857 & 0.5899 \\
 & MIF-$\Delta$logit & 0.1585 & 0.1166 & 0.2918 & 0.2192 & 1.9092 & 1.3301 & 0.5749 \\
 & RDE-Linear & 0.2903 & 0.2632 & 0.4185 & 0.3514 & 1.7832 & 1.2159 & 0.6059 \\
 & BA-Cycle & 0.3722 & 0.3201 & 0.4552 & 0.4097 & 1.8402 & 1.3026 & 0.6657 \\ \midrule
 \multirow{13}{*}{Supervised} & DDGPred & 0.3750 & 0.3407 & 0.6580 & 0.4687 & {1.4998} & {1.0821} & 0.6992 \\
 & End-to-End & 0.3873 & 0.3587 & 0.6373 & 0.4882 & 1.6198 & 1.1761 & 0.7172 \\
 & MIF-Network & 0.3965 & 0.3509 & 0.6523 & 0.5134 & 1.5932 & 1.1469 & 0.7329 \\ \cmidrule{2-9}
 & \multirow{2}{*}{RDE-Network} & {0.4448} & {0.4010} & 0.6447 & {0.5584} & 1.5799 & 1.1123 & {0.7454} \\
 & & {0.4422} & {0.3914} & 0.6386 & 0.5340 & 1.5904 & 1.1307 & 0.7380 \\ \cmidrule{2-9}
 &\multirow{2}{*}{DiffAffinity} & 0.4220 & 0.3970 & {0.6690} & 0.5560 & 1.5350 & 1.0930 & 0.7440 \\ 
 & & 0.4040 & 0.3719 & {0.6738} & 0.5508 & 1.5835 & 1.1740 & {0.7568}\\ \cmidrule{2-9}
 & \multirow{2}{*}{Prompt-DDG} & {0.4712} & {0.4257} & {0.6772} & {0.5910} & {1.5207} & {1.0770} & {0.7568} \\
 & & {0.4369} & {0.3904} & \underline{0.6661} & {0.5626} & \underline{1.5415} & \underline{1.0989} & {0.7369} \\ \cmidrule{2-9}
 & \multirow{2}{*}{ProMIM} & 0.4640 & 0.4310  & {0.6720} & {0.5730} & 1.5160 & 1.0890 & 0.7600 \\
 & & 0.4085 & 0.3846 & 0.6376 & 0.5357 & 1.6403 & 1.2022 & 0.7480 \\ \cmidrule{2-9}
 & \multirow{2}{*}{BA-DDG} & \underline{0.5453} & \underline{0.5134} & \underline{0.7118} & \underline{0.6346} & \underline{1.4516} & \underline{1.0151} & \underline{0.7726} \\ 
 & & \underline{0.5266} & \underline{0.5040} & {0.6469} & \underline{0.6034} & 1.6650 & 1.1942 & \underline{0.7533} \\ \midrule
  \cellcolor{lm_purple}
  & \cellcolor{lm_purple} 
  & \cellcolor{lm_purple}\textbf{0.5681} 
  & \cellcolor{lm_purple}\textbf{0.5184}
  & \cellcolor{lm_purple}\textbf{0.7385}
  & \cellcolor{lm_purple}\textbf{0.6516}
  & \cellcolor{lm_purple}\textbf{1.3901}
  & \cellcolor{lm_purple}\textbf{0.9871}
  & \cellcolor{lm_purple}\textbf{0.7941}
\\
\multirow{-2}{*}{\cellcolor{lm_purple}Ours} 
  & \multirow{-2}{*}{\cellcolor{lm_purple}EBM-DDG}
  & \cellcolor{lm_purple}\textbf{0.5472} 
  & \cellcolor{lm_purple}\textbf{0.5076}
  & \cellcolor{lm_purple}\textbf{0.7140}
  & \cellcolor{lm_purple}\textbf{0.6526}
  & \cellcolor{lm_purple}\textbf{1.4436}
  & \cellcolor{lm_purple}\textbf{1.0102}
  & \cellcolor{lm_purple}\textbf{0.7866}
\\
 \bottomrule
\end{tabular}}
\end{center}
\end{table*} 

\noindent\textbf{Comparison with Baselines.}
In~\cref{tab:1}, we see that the core deep learning baselines outperform all other $\ddg$ prediction methods. Notably, EBM-DDG outperforms all deep learning-based methods and presents a new state-of-the-art. EBM-DDG improves on the majority of baselines by a wide margin, achieving per-structure Pearson and Spearman correlations of 0.5681 and 0.5184 compared to Prompt-DDG's 0.4712 and 0.4257, for example in overall mutations. Importantly, EBM-DDG clearly improves on BA-DDG, the most competitive baseline, by margins similar to how BA-DDG improves on previous baselines metrics like AUROC, RMSE, and MAE. These trends persist for multiple-point mutation performance as reported in \cref{sec:multi}, which is known to be challenging due to the complexity of interacting mutations.
In~\cref{fig:corr}, we visualize EBM-DDG's per-structure correlation performance against competing baseline models under our evaluation setting using 10\% of each training fold as a validation set. EBM-DDG's mean correlation performance is superior to all baselines, most notably BA-DDG. In both correlation metrics, we also see that EBM-DDG exhibits smaller variance, showcasing the benefit of having a strong inductive bias.
\begin{figure*}[!htbp]
    \centering
    \includegraphics[width=0.85\linewidth]{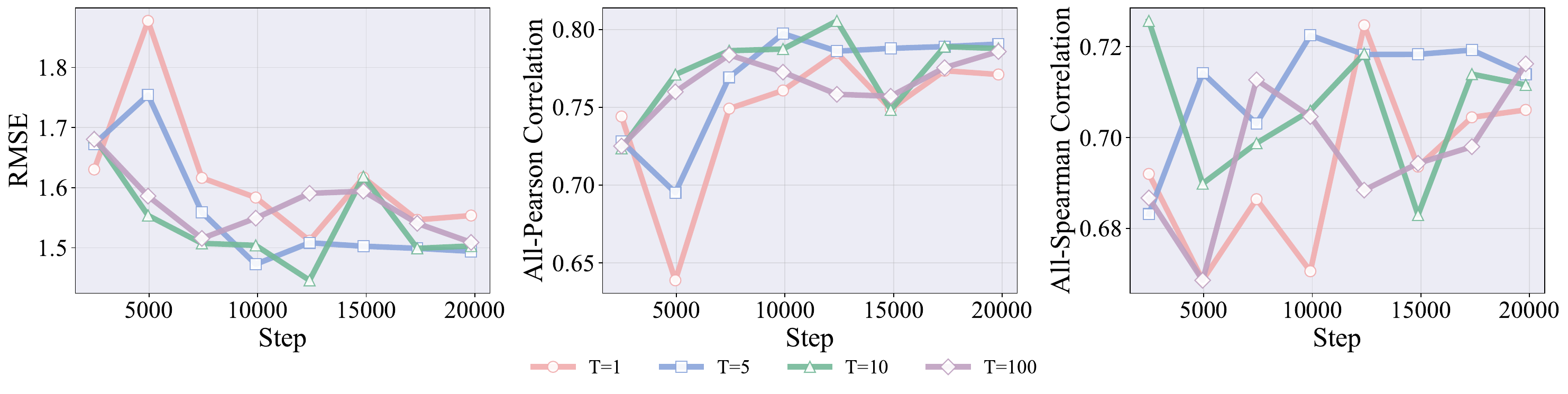}
    \caption{RMSE, All-Pearson, and All-Spearman correlation coefficients over time as the number of denoising steps $T \in \{1, 5, 10, 100\}$ varies. We see that, generally, denoising with $T=10$ is enough to achieve optimal performance across the board.}
    \Description[Too few denoising steps hurts performance]{Overall, denoising with few steps hurts RMSE and all-structure correlation performance. T=10 yields a balance between performance and efficiency.}
    \label{fig:T}
\end{figure*}
\subsection{Antibody Optimization -- SARS-CoV-2}
\label{sec:antibody}
\begin{table}[!tbp]
    \begin{center}
        \caption{Rankings of the five favorable mutations on the antibody against SARS-CoV-2 by various state-of-the-art models. DiffAffinity, ProMIM, RDE-Network, Prompt-DDG, used checkpoints finetuned on SKEMPI using our train/valid/test split.}
        \label{tab:2}
        \resizebox{\columnwidth}{!}{
        \begin{tabular}{l|ccccc|c}
        \midrule
        \textbf{Method} & TH31W & AH53F & NH57L & RH103M & LH104F & Average \\ \midrule
        Rosetta & 10.73\% & 76.72\% & 93.93\% & 11.34\% & 27.94\% & 44.13\% \\
        FoldX & 13.56\% & 6.88\% & \textbf{5.67\%} & 16.60\% & 66.19\% & 21.78\% \\ \midrule
        DDGPred & 68.22\% & 2.63\% & 12.35\% & \textbf{8.30\%} & 8.50\% & 20.00\% \\
        End-to-End & 29.96\% & \underline{2.02\%} & 14.17\% & 52.43\% & 17.21\% & 23.16\% \\ \midrule
        DiffAffinity & \textbf{2.43\%} & 26.52\% & 13.16\% & 89.27\% & 12.75\% & 32.84\% \\
        ProMIM & 7.29\% & 34.82\% & 42.11\% & 90.69\% & \textbf{1.01\%} & 43.72\% \\
        RDE-Network & \underline{4.25\%} & 3.64\% & 11.94\% & 75.51\% & \underline{2.23\%} & 23.84\% \\
        Prompt-DDG & 5.26\% & \textbf{0.20\%} & 44.13\% & 31.38\% & 76.32\% & 20.24\% \\
        BA-DDG & 14.17\% & 11.34\% & \underline{6.07}\% & 15.79\% & 17.00\% & \underline{12.87\%} \\ \midrule
        \rowcolor{lm_purple}
        EBM-DDG & 12.96\% & 24.70\% & {10.73\%} & \underline{8.91\%} & \underline{2.23}\% & \textbf{11.90\%} \\ 
        \bottomrule        \end{tabular}} \vspace{-1em}
    \end{center}
\end{table} 
$\ddg$ prediction is highly useful in scenarios for identifying desirable mutations, particularly those with high binding affinity, among the vast space of possible mutations. In this subsection, we analyze EBM-DDG's ability to rank mutations for antibody optimization against SARS-CoV-2, following an experimental setting by~\citep{shan2022ddgpred} as a case study. In this experiment, we take models trained on SKEMPI v2.0 under our train/validation/test split and task them with predicting the $\ddg$ values of 494 possible single-point mutations at the 26 sites within the CDR region of the antibody heavy chain. We then rank these predictions in descending order. Higher-ranked mutations (lower $\ddg$) indicate higher antibody effectiveness due to the added stability of lowered binding free energy. We measure performance using the ranking of each mutation divided by the number of total possible mutations, represented as a percentage. ~\cref{tab:2} shows how EBM-DDG compares with the strongest deep learning baselines DiffAfinity, ProMIM, RDE-Network, Prompt-/DDG, and BA-DDG. We also compare with classical methods such as Rosetta, FoldX as well as DDGPred and the End-To-End model by~\citep{luo2023rotamer} as preliminary deep learning baselines. In this experiment, we see that EBM-DDG has the highest average rank percentage of 11.90\% compared to BA-DDG's 12.87\% and FoldX's 21.78\%. We also observe that EBM-DDG is one of the only methods besides RDE-Network to achieve second-best metrics in more than one mutation with other methods achieving first- or second-best metrics in only one mutation, highlighting the strength of the structure-aware inductive bias built into EBM-DDG. A conjecture as to why EBM-DDG may still struggle with certain mutations is that this task is based on predicting mutational effects on CDRs, which may involve ``dangling'' residues, or subregions of proteins that are geometrically disjoint from its neighbors, posing a challenge for the denoising process of DSMBind which is pre-trained on comprehensive crystal structures and may give unrealistic estimates for non-contiguous residues.
\subsection{Case Study -- Structurally Flexible Mutations}
\label{sec:flex}
\begin{table}[!tbp]
    \begin{center}
        \caption{$\ddg$ RMSE and MAE for mutations to glycine/proline.}
        \label{tab:flex}
        \resizebox{0.78\columnwidth}{!}{
        \begin{tabular}{c|cc|cc}
        \toprule
        \multirow{2}{*}{\textbf{Method}} & \multicolumn{2}{c|}{To Glycine} & \multicolumn{2}{c}{To Proline} \\
         & RMSE & MAE & RMSE & MAE \\
        \midrule
        ProMIM      & 1.524 & 1.348 & 1.264 & 1.041 \\
        Prompt-DDG  & 1.864 & 1.651 & 1.289 & 1.104  \\
        BA-DDG      & 1.307 & 1.254 & 1.354 & 1.059 \\ \midrule
        \rowcolor{lm_purple}
        EBM-DDG     & \textbf{1.110} &\textbf{0.941} &  \textbf{1.175} & \textbf{0.960}  \\
        \bottomrule
        \end{tabular}}
        \vspace{-1em}
    \end{center}
\end{table}

In this subsection, we analyze EBM-DDG's performance on mutations to glycine and proline: residues known for their structural flexibility and inflexibility, respectively. This acts as a test of EBM-DDG's ability to model mutations that are more likely to affect backbone structure. Due to the relatively low sample sizes of these kinds of mutations (11 and 16), we report RMSE and MAE rather than correlation coefficients which can be misleading in low-data settings. We compare with ProMIM~\citep{mo2024multi}, a method pre-trained on backbone contact maps, and Prompt-DDG~\citep{wu2024prompt}, a baseline pre-trained on autoencoding backbone coordinates. These methods stand as representative baselines that take backbone structure into account, although they do not conduct any sampling during inference, unlike EBM-DDG. We also compare with BA-DDG to verify that EBM-DDG's energy-based inductive bias can account for structural effects of mutations. We see in~\cref{tab:flex} that EBM-DDG consistently outperforms all of these baselines, achieving the lowest RMSE and MAE values on glycine and proline mutations. This validates EBM-DDG's energy-based inductive bias that enables directly sampling backbone coordinates, allowing for more accurate modeling of backbone-sensitive mutations.
\subsection{Sensitivity to Number of Denoising Steps}
\label{sec:T}
A key question is how sensitive EBM-DDG's performance is with respect to the generative process by which mutant's coordinates are sampled. A key hyperparameter governing this process is $T$: the number of denoising steps used during sampling according to Langevin dynamics. In this subsection, we analyze the effect of varying $T$ on three key performance metrics: RMSE, Per-Pearson coefficient, and Per-Spearman coefficient. Specifically, we vary $T \in \{1, 5, 10, 100\}$ and plot performance across 20K optimization steps in~\cref{fig:T} on one of our CV folds.
Generally, using $T=1$ is the worst performer with the highest RMSE and lowest All-Pearson and All-Spearman coefficients overall. This is expected given that sampling with Langevin dynamics typically requires several steps before reaching anything resembling the target distribution~\citep{purohit2024posteriorsamplinglangevindynamics,kirchmeyer2024scorebased}. Next, we see that $T = 100$ results in initially superior performance in earlier optimization steps for RMSE and All-Pearson performance but quickly lags behind, indicating drift towards unrealistic coordinates. We also observed a substantial decrease in efficiency with $T=10$ requiring on average 12.1 minutes per epoch, while $T = 100$ requires on average 50.8 minutes per epoch, indicating the need to find a more efficient setting for $T$. Finally, we see that $T = 10$ yields the best overall results with $T = 5$ coming close behind, granting a balance between fidelity and efficiency. In line with this, other works~\citep{gnaneshwar2022scorebasedgenerativemodelsmolecule} find that $T = 10$ is enough for sampling biomolecules from score-based generative models.
\section{Conclusion}
To conclude, this work introduces EBM-DDG, an energy-based approach to $\ddg$ prediction that more comprehensively considers the interplay between structure and sequence, leveraging insights from statistical mechanics to simplify the estimation of complex probabilities for a stronger inductive bias. For future work, we note that DSMBind is only one of many energy functions to choose from. For example, MACE~\citep{Batatia2022mace} and Allegro~\citep{Musaelian2022-sf} are energy functions designed for molecular dynamics/material science applications, but their performance on modeling protein-protein interactions is unknown. It would be interesting to see how other neural potentials can contribute to this approach to $\ddg$ prediction. We also want to explore other ways to relax hidden assumptions, such as assuming a representative structure given a sequence. Expanding this method to IDPs or membrane proteins would be of interest. Finally, we highlight the need for stronger evaluation in the $\ddg$ prediction literature beyond a single cross-validation set from SKEMPI v2.0.

\begin{acks}
This work is supported in part by the National Science Foundation (NSF) under grants IIS-2006844, IIS-2144209, IIS-2223769, CNS-2154962, BCS-2228534, and CMMI-2411248; the Office of Naval Research (ONR) under grant N000142412636; the Commonwealth Cyber Initiative (CCI) under grant VV-1Q24-011; and the UVA SEAS Research Innovation Award.
\end{acks}


\bibliographystyle{ACM-Reference-Format}
\balance
\bibliography{sample-base}

\appendix

\section{Multi-Point Mutation Results}
\label{sec:multi}
We include multi-point mutation results in addition to the overall results on SKEMPI v2.0 under the original train/test splits of~\citep{liu2023predicting}. We include each method's reported performance from their respective papers. Given their availability, we include the performance of PPIformer~\citep{bushuiev2024learning} and Refine-PPI~\citep{wu2025dynamicsinspired} in \cref{tab:skempi_single_multi_results}. We see that in nearly all metrics for both single- and multiple-point mutations, EBM-DDG outperforms all other baselines. 
\begin{table*}[!htbp]
    \caption{Performance breakdown between all-, single-, and multi-point mutations on SKEMPI v2.0. Best metrics are bolded, and second-best metrics are underlined.}
    \centering
    \resizebox{0.88\textwidth}{!}{
        \begin{tabular}{llccccccc}
            \toprule
            \multirow{2}{*}{\textbf{Method}} & \multirow{2}{*}{\textbf{Mutations}} & \multicolumn{2}{c}{\textbf{Per-Structure}} & \multicolumn{5}{c}{\textbf{Overall}} \\
            \cmidrule(lr){3-4} \cmidrule(lr){5-9}
            & & \textbf{Pearson $\uparrow$} & \textbf{Spear. $\uparrow$} & \textbf{Pearson $\uparrow$} & \textbf{Spear. $\uparrow$} & \textbf{RMSE $\downarrow$} & \textbf{MAE $\downarrow$} & \textbf{AUROC $\uparrow$} \\
            \midrule
            \multirow{3}{*}{Rosetta}
            & all & 0.3284 & 0.2988 & 0.3113 & 0.3468 & 1.6173 & 1.1311 & 0.6562 \\
            & single & 0.3510  & 0.4180 &  0.3250  & 0.3670 &  1.1830  & 0.9870  & 0.6740 \\
            & multiple & 0.1910  & 0.0830  & 0.1990 &  0.2300  & 2.6580  & 2.0240  & 0.6210 \\
            \midrule
            \multirow{3}{*}{FoldX}
            & all & 0.3789 & 0.3693 & 0.3120 & 0.4071 & 1.9080 & 1.3089 & 0.6582 \\
            & single & 0.3820 &  0.3600  & 0.3150 &  0.3610  & 1.6510  & 1.1460  & 0.6570 \\
            & multiple & 0.3330  & 0.3400  & 0.2560 &  0.4180  & 2.6080  & 1.9260 &  0.7040 \\
            \midrule
            \multirow{3}{*}{DDGPred}
            & all & 0.3750 & 0.3407 & 0.6580 & 0.4687 & 1.4998 & 1.0821 & 0.6992 \\
            & single & 0.3711  & 0.3427  & 0.6515  & 0.4390  & 1.3285  & 0.9618 &  0.6858 \\
            & multiple & 0.3912 &  0.3896 &  0.5938  & 0.5150  & 2.1813 &  1.6699 &  0.7590 \\
            \midrule
            \multirow{3}{*}{End-to-End}
            & all & 0.3873 & 0.3587 & 0.6373 & 0.4882 & 1.6198 & 1.1761 & 0.7172 \\
            & single & 0.3818  & 0.3426  & 0.6605  & 0.4594  & 1.3148  & 0.9569  & 0.7019 \\
            & multiple & 0.4178  & 0.4034  & 0.5858  & 0.4942 &  2.1971  & 1.7087 &  0.7532 \\
            \midrule
            \multirow{3}{*}{RDE-Network}
            & all & 0.4448 & 0.4010 & 0.6447 & 0.5584 & 1.5799 & 1.1123 & 0.7454 \\
            & single & 0.4687 &  0.4333  & 0.6421 &  0.5271 &  1.3333  & 0.9392  & 0.7367 \\
            & multiple & 0.4233 &  0.3926 &  0.6288  & 0.5900 &  2.0980  & 1.5747 &  0.7749\\
            \midrule
            \multirow{3}{*}{DiffAffinity}
            & all & 0.4220 & 0.3970 & 0.6609 & 0.5560 & 1.5350 & 1.0930 & 0.7440 \\
            & single & 0.4290  & 0.4090  & 0.6720  & 0.5230 &  1.2880  & 0.9230 & 0.7330\\
            & multiple & 0.4140 &  0.3870 &  0.6500 &  0.6020 &  2.0510 &  1.5400 & 0.7840\\
            \midrule
            \multirow{3}{*}{Prompt-DDG}
            & all & 0.4712 & 0.4257 & 0.6772 & 0.5910 & 1.5207 & 1.0770 & 0.7568 \\
            & single & 0.4736 &  0.4392 &  0.6596  & 0.5450  & 1.3072  & 0.9191 &  0.7355 \\
            & multiple & 0.4448 &  0.3961  & \underline{0.6780}  & \underline{0.6433}  & 1.9831  & \underline{1.4837} & 0.8187 \\
            \midrule
            \multirow{3}{*}{ProMIM}
            & all & 0.4640 & 0.4310 & 0.6720 & 0.5730 & 1.5160 & 1.0890 & 0.7600 \\
            & single & 0.4660 &  0.4390  & 0.6680 &  0.5340 &  1.2790 &  0.9240 &  0.7380 \\
            & multiple & 0.4580  & 0.4250  & 0.6660 &  0.6140 &  \underline{1.9630}  & 1.4910 &  \underline{0.8250}\\
            \midrule
            \multirow{3}{*}{PPIFormer}
            & all & 0.4281 & 0.3995 & 0.6450 & 0.5304 & 1.6420 & 1.1186 & 0.7380 \\
            & single & 0.4192 & 0.3796 & 0.6287 & 0.4772 & 1.4232 & 0.9562 & 0.7213 \\
            & multiple & 0.3985 & 0.3925 & 0.6405 & 0.5946 & 2.1407 & 1.5753 & 0.7893 \\ \midrule
            \multirow{3}{*}{Refine-PPI}
            & all & 0.4561 & 0.4374 & 0.6592 & 0.5608 & 1.5643 & 1.1093 & 0.7542 \\
            & single & 0.4701 & 0.4459 & 0.6658 & 0.5153 & 1.2978 & 0.9287 & 0.7481 \\
            & multiple & 0.4558 & 0.4289 & 0.6458 & 0.6091 & 2.0601 & 1.554 & 0.8064 \\ 
            \midrule
            \multirow{3}{*}{BA-DDG}
            & all & \underline{0.5453} & \underline{0.5134} & \underline{0.7118} & \underline{0.6346} & \underline{1.4516} & \underline{1.0151} & \underline{0.7726}  \\
            & single & \underline{0.5606} & \underline{0.5192} & \underline{0.7321} & \underline{0.6157} & \underline{1.1848} & \underline{0.8409} & \underline{0.7662}  \\
            & multiple & \underline{0.4924} & \textbf{0.4959} & {0.6650} & {0.6293} & 2.0151 & 1.4944 & 0.7875  \\
            \midrule
            \cellcolor{lm_purple}                                        
             & \cellcolor{lm_purple}all
             & \cellcolor{lm_purple}\textbf{0.5681}
             & \cellcolor{lm_purple}\textbf{0.5184}
             & \cellcolor{lm_purple}\textbf{0.7385}
             & \cellcolor{lm_purple}\textbf{0.6516}
             & \cellcolor{lm_purple}\textbf{1.3901}
             & \cellcolor{lm_purple}\textbf{0.9871}
             & \cellcolor{lm_purple}\textbf{0.7941}
            \\
            \cellcolor{lm_purple}                                        
             & \cellcolor{lm_purple}single
             & \cellcolor{lm_purple}\textbf{0.5693}
             & \cellcolor{lm_purple}\textbf{0.5290}
             & \cellcolor{lm_purple}\textbf{0.7343}
             & \cellcolor{lm_purple}\textbf{0.6160}
             & \cellcolor{lm_purple}\textbf{1.1754}
             & \cellcolor{lm_purple}\textbf{0.8302}
             & \cellcolor{lm_purple}\textbf{0.7775}
            \\
            \multirow{-3}{*}{\cellcolor{lm_purple}EBM‑DDG}               
             & \cellcolor{lm_purple}multiple
             & \cellcolor{lm_purple}\textbf{0.5361}
             & \cellcolor{lm_purple}\underline{0.4873}
             & \cellcolor{lm_purple}\textbf{0.7236}
             & \cellcolor{lm_purple}\textbf{0.6965}
             & \cellcolor{lm_purple}\textbf{1.8628}
             & \cellcolor{lm_purple}\textbf{1.4206}
             & \cellcolor{lm_purple}\textbf{0.8420}
            \\
            \bottomrule
        \end{tabular}
    }
    \label{tab:skempi_single_multi_results}
\end{table*}
\section{ProteinMPNN}
\label{sec:proteinmpnn}
ProteinMPNN is a protein inverse folding model trained to predict amino acid sequences given a 3D structure. Following~\citep{dauparas2022robust}, ProteinMPNN is pre-trained on 19.7k monomers from the PDB based on the CATH4.2 40\% non-redundant set of proteins~\citep{Orengo1997-kg} using half-precision (16-bit) gradients. For our $\ddg$ prediction, we only need the log-probabilities of mutated residues, not the entire sequence. Therefore, following~\citep{jiao2024boltzmann,dutton2024improving}, we designate mutation sites $S_D$ and fix the rest of the sequence $S$, which we call sequence context in~\cref{fig:pipeline}. Borrowing notation and equations from~\citep{jiao2024boltzmann}, we randomly sample a decoding order $\pi = (s_1, s_2, \dots, s_n)$, and expand $P(S \mid X)$ into the product
\begin{align*}
    P(S \mid X) &= P(S_D \mid X \setminus S_D) \\
    &= P(S_D \setminus \{s_1\} \mid X, S \setminus S_D \cup \{s_1\}) \cdot P(\{s_1\} \mid X, S \setminus S_D) \\
    &= \cdots
\end{align*}
and so-on. 

\section{Sampling Details}
\label{sec:sampling}
\subsection{Energy Function}
The approach to use a learned energy function under the denoising score matching objective is from~\citep{jin2023dsmbind,jin2023unsupervised}. We adapt their approach by swapping their SRU++-based frame-averaging encoder~\citep{lei2021srupp,puny2022frame} with a 1-layer Invariant Point Attention~\citep{Jumper2021-hp} encoder without pair representations with a hidden dimension of 256. Following~\citep{jin2023dsmbind}, we initialize our energy function with 2560-dimensional embeddings from ESM-2 3B~\citep{lin2022language} for each residue and reduce the dimension down to 256 via a linear layer. The all-atom interaction block referenced in~\cref{fig:pipeline} is composed of two 1-layer MLPs applied to the binder and target chains, respectively, followed by a summation over all pairwise products of each atom embedding. Specifically, given binder and target embeddings $H_\text{bnd} \in \mathbb{R}^{m \times d}$ and $H_\text{tgt} \in \mathbb{R}^{n \times d}$, we use 
$$\mathcal E_\theta\left(H_{\text{bnd}}, H_{\text{tgt}}\right) = \sum_{i, j}^n\left(\operatorname{MLP}_{\text{bnd}}\left(H_\text{bnd}\right)\operatorname{MLP}_{\text{tgt}}\left(H_\text{tgt}\right)^\top\right)_{i,j}$$
as our learned energy function.

\subsection{Invariance of Score-Matching Minima}
Here, we clarify how the minima of the denoising score-matching objective are equal up to a scaling and shifting of the energy term $\mathcal E$. Assuming Gaussian noise $\epsilon \sim N(0, \sigma^2I)$, the denoising score-matching objective is $L = \mathbb E\left[||\nabla_{\hat{X}} \mathcal E(\hat X)-\epsilon/\sigma^2||^2\right]$. 
Suppose we shift the learned score function by a constant $c$. Then $$L_{shift}=\mathbb E\left[||\nabla_{\hat{X}}(\mathcal E(\hat X)+c)-\epsilon/\sigma^2||^2\right]=\mathbb E\left[||\nabla_{\hat{X}}\mathcal E(\hat X)-\epsilon/\sigma^2||^2\right]$$ since the gradient cancels the added constant. When scaling by $c$, we have $$L_{scale}=\mathbb E\left[||\nabla_{\hat{X}}(c\mathcal E(\hat X))-\epsilon/\sigma^2||^2\right]=c^2\mathbb E\left[||\nabla_{\hat{X}}\mathcal E(\hat X)-\epsilon/c\sigma^2||^2\right],$$ which achieves its minimum when $\nabla_{\hat{X}} \mathcal E(\hat X) = \epsilon/c\sigma^2$, or when the learned score is a scaled version of the true score. While the scale of the gradient changes, our objective's optima have not changed.

\subsection{Langevin Dynamics}
Denoising score-matching is a popular approach for generative modeling inspired by~\citep{song2019generative,hyvarinen2005estimation} that seeks to model the gradient of the log-density (i.e., the score) $\nabla_x \log p(x)$ of the data distribution from noise-perturbed samples. Typically, a neural network with parameters $\theta$ is used to estimate this score $s_\theta \approx \nabla_x \log p(x)$. A standard way of sampling from such models is via Langevin dynamics~\citep{welling2011bayesian}: a process by which a random sample is iteratively moved towards high-density regions of the data distribution following the learned score function. Specifically, given a random sample $x_0 \sim P(X)$ initialized from some noise prior $P$, one iterates over $T$ steps, each time updating $$x_t = x_{t-1} + \alpha s_\theta(x_{t-1}) + \eta\epsilon$$ where $\alpha$ is a step size that controls the score, $\eta$ is a step size controlling the amount of noise being sampled, and $\epsilon \sim \mathcal N(0, 1)$. $\alpha$ and $\eta$ may be dynamic according to some time-dependent schedule in order to facilitate exploration of high-density regions~\citep{song2019generative}. In our case, we use Langevin dynamics to sample the backbone coordinates of mutated residues where the score of the noise distribution is given by $ \log p(\epsilon) = - \epsilon /\sigma^2$ where $\epsilon \sim \mathcal N(0, 0.1{I})$. We use a cosine annealing schedule during denoising following~\citep{nichol2021improved} with an initial $\alpha_0 = 0.001, \eta_0 = 0.01$.

\section{Example Conformations}
\begin{figure}
    \centering
    {
        \includegraphics[width=0.58\linewidth]{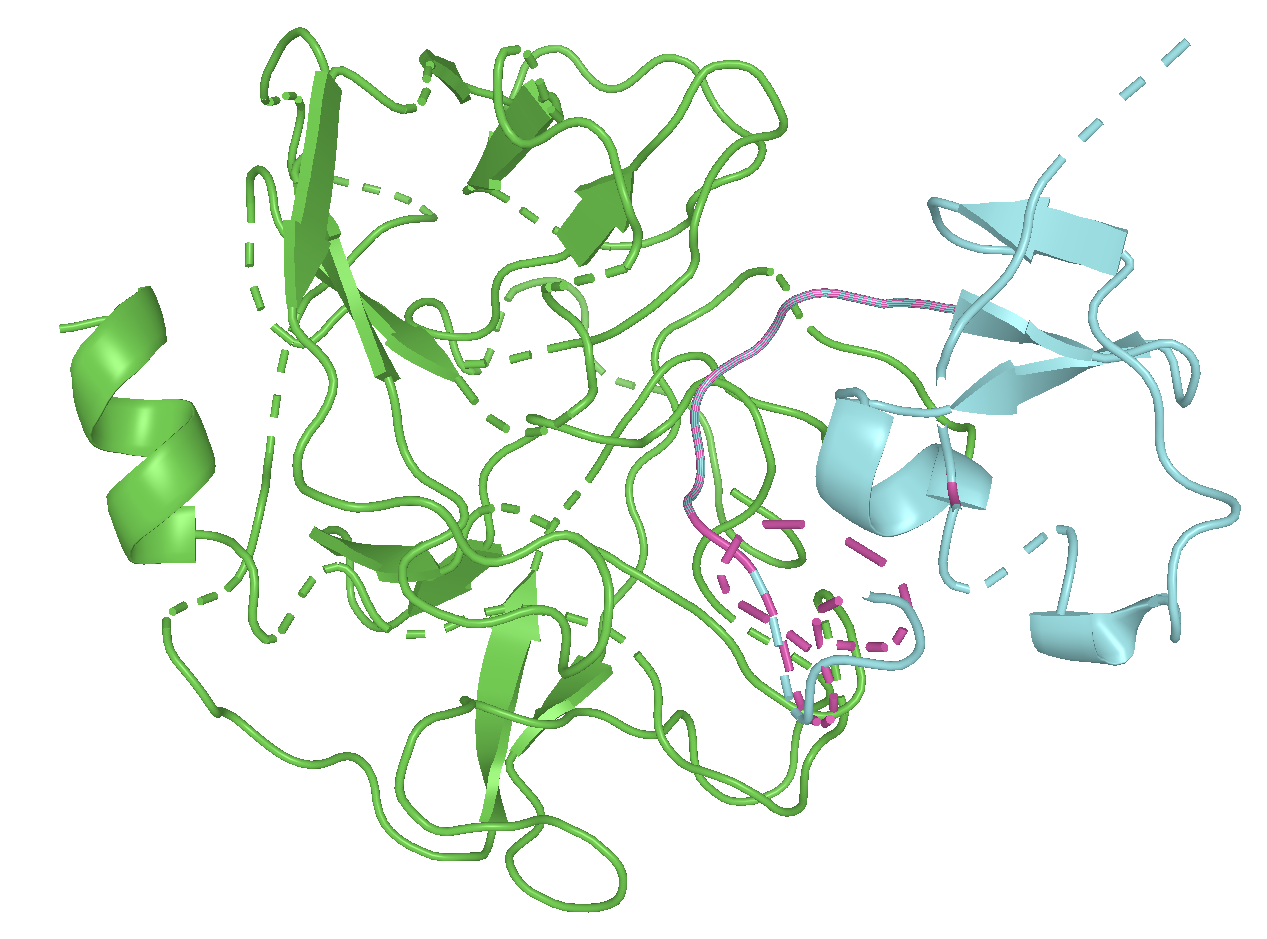}
        \includegraphics[width=0.4\linewidth]{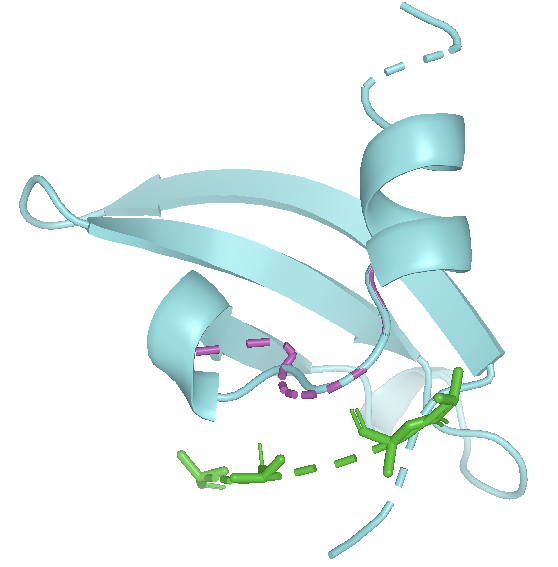}
    }
    \caption{(Left) Visualization of the mutation PI14G in 1PPF (RMSE=0.4045). (Right) Visualization of DA40F in 1KNE (RMSE=0.0020). Experimental wild-type structures are in cyan/green, and sample structures are in magenta.}
    \Description[Generated mutant structures deviate from wild-type coordinates]{We give examples of mutant structures with relatively low RMSE, specifically PI14G in 1PPF (RMSE=0.4045) and DA40F in 1KNE (RMSE=0.0020). As expected, mutating amino acids with disparate masses induces larger conformational changes.}
    \label{fig:viz}
\end{figure}
In \cref{fig:viz}, we show some example sampled conformations from our learned energy model. We see that, for mutations like $P\rightarrow G$ on the left, the backbone is perturbed substantially. On the right, we see similar phenomena when converting from a light amino acid like aspartic acid (D) to a heavier amino acid like phenylalanine (F) with the disruption of an $\alpha$-helix. It is important to note that these samples are difficult to evaluate without experimentally validated mutant structures.

\end{document}